# What's in a Measurement? Using GPT-3 on SemEval 2021 Task 8 - MeasEval

Curt Kohler, Ron Daniel Jr.
Elsevier Labs

**Abstract**

In the summer of 2020 OpenAI released its GPT-3 autoregressive language model to much fanfare. While the model has shown promise on tasks in several areas, it has not always been clear when the results were cherry-picked or when they were the unvarnished output. We were particularly interested in what benefits GPT-3 could bring to the SemEval 2021 MeasEval task - identifying measurements and their associated attributes in scientific literature. We had already experimented with multi-turn questions answering as a solution to this task. We wanted to see if we could use GPT-3's few-shot learning capabilities to more easily develop a solution that would have better performance than our prior work. Unfortunately, we have not been successful in that effort. This paper discusses the approach we used, challenges we encountered, and results we observed. Some of the problems we encountered were simply due to the state of the art. For example, the limits on the size of the prompt and answer limited the amount of the training signal that could be offered. Others are more fundamental. We are unaware of generative models that excel in retaining factual information. Also, the impact of changes in the prompts is unpredictable, making it hard to reliably improve performance.

**Background**

There has been a virtual arms race in NLP development as contenders compete for the SOTA throne by creating larger and larger models. In the summer of 2020 OpenAI released its GPT-3 (Generative Pre-Trained Transformer - 3) autoregressive NLP language model to much fanfare[1]. At 175 billion parameters, it represented an order of magnitude increase over its nearest competitor, Microsoft's Turing-NLG model[2].

| Year | Model | Parameter size |
|---|---|---|
| **2018** | ELMo | 94M |
| **2018** | GPT | 110M |
| **2018** | BERT-Large | 340M |
| **2019** | Transformer ELMo | 465M |
| **2019** | GPT-2 | 1.5B |
| **2019** | Megatron-LM | 8.3B |
| **2020** | Turing-NLG | 17B |
| **2020** | GPT-3 | 175B |

Previous language models like BERT have relied on pretraining on a large corpus followed by custom training data to fine-tune the base model to a specific task. The GPT-X family of models were presented as sufficiently generic and pretrained on enough data that they did not require extensive fine tuning. In fact, the models were explicitly presented as being "few-shot" learners. This implies that for many novel tasks, GPT models could be fed only a few supervised example text sequences and they could distill enough intent to complete the task on the target data input. If accurate, GPT-3 represents an attractive alternative for dealing with tasks where extensive training data or the significant computational resources needed for fine tuning are unavailable. GPT-3 has been tested on a variety of different tasks including language translation, code generation, writing creative fiction and short news stories, and named entity resolution to name a few[1][3][4][5][6]. In some cases, GPT-3 exceeded SOTA, and in others the results met or lagged published benchmarks.

For our evaluation of GPT-3, we decided to apply it to one of the tasks from SemEval 2021 and compare the results to other submissions for that task. Specifically, we chose Task 8 - MeasEval - Counts and Measures[7]. The MeasEval task involves identifying measurements and their associated attributes as they occur within scientific literature. While this may sound like a trivial task, these entities and attributes related to a measurement are frequently presented in ambiguous and varied ways between various disciplines, between authors within the same discipline, and even by the same author within the same paper.

The MeasEval task is composed of five subtasks where a system processing scientific text paragraphs must identify:

- All measured Quantities present in the paragraph with their associated spans
- The Unit of measurement for each quantity, if present, and any modifiers to the quantity (approximate value, range of values, etc.)
- The Entity and/or Property of the Entity being measured, as well as their associated spans in the paragraph
- Any Qualifiers needed to represent context of the quantity along with their associated spans, such as the pressure at which a temperature was measured.
- The Relationships between the identified Quantities, Entities, Properties, and Quantities

The MeasEval data consists of 248 training paragraphs with 2,531 hand-crafted annotations of the various entities, attributes, and relationships plus an evaluation set of 135 paragraphs containing 1,490 annotations. The annotations associated with the evaluation set were released after the official MeasEval task challenge period ended. Additionally, the task also provides a standalone, Python scoring tool that can be used to evaluate a test run's results against the evaluation gold set. All of these items are available for download on the MeasEval GitHub site[8].

While OpenAI has not made the actual GPT-3 models public, they have created a restricted-access beta website where registered users may submit requests against GPT-3 via a simple web interface or leverage a Python based API. Either approach exposes the configuration parameters that control various aspects of the model's processing. One of the configuration settings we leveraged was the GPT-3 "Temperature" or degree of randomness of the generated response. Another parameter we experimented with was the specific engine to use. GPT-3 has 4 distinct engines: davinci, curie, babbage, and ada. These represent different sizes, levels of complexity, and intended uses. Registered beta users were provided a set number of credits which were consumed as requests were processed. Credits were consumed at varying rates based on the size of the submitted prompt/response and the engine used to process them.

## Experimental Procedure

### *Initial Investigations*

We received our beta site credentials and an initial allocation of credits. While the provided credits would be adequate for playing with a few scenarios they would be tight for conducting a thorough evaluation of many different alternatives over all the MeasEval data. Our initial experiments varied settings for "temperature", GPT-3 model, number of few-shot examples, and output format with the goal of obtaining a configuration that would be used in our measured attempt at the MeasEval test. During these experiments we observed:

- the davinci engine generated slightly more accurate output than the curie (or other) engines for this task
- even low "temperature" settings introduced randomness into the generated text for identical submissions - including text spans not contained within the target sentence. Those hallucinations of new text were not suitable for our application, so we set the temperature to 0 in all later work.
- The number, and the order, of the few-shot examples significantly impacted the output. Additional examples did not always lead to better results and swapping example order could change the results.
- The models had problems learning to generate required output fields like span offsets and unique identifiers. Frequently it would output a repetitive or a random value.
- The models had issues identifying Quantity modifiers, Relationship modifiers, and Qualifier entities.
- The model sometimes would begin generating a repeating textual string until the max length was reached.

Initially we tried to have the system produce output that matched a simplified version of the tab separated value (TSV) format of the MeasEval annotations, as shown in Figure 1. As noted above, GPT-3 was too unreliable when generating several the items present in the simplified format. We hypothesized that the various Quantity modifiers were problematic due to the variety of values that could be present relative to the training samples presented. We suspected that the Relationship modifiers were faulty because if the underlying issues GPT-3 appeared to have while generating unique ids for each item. We also hypothesized that the detailed structure of the task was just too much to learn given only a few shots in lieu of more comprehensive fine tuning. We further simplified the desired output format by removing problematic aspects (ids, offsets, Entity modifiers, and Qualifier Entities) to create an even more simplified output format as shown in Figure 2. When possible, these dropped field values would be re-created by post-processing the GPT-3 results.

```
Text: The averaged power extracted during one cycle increased by 22% from 48.4 MW
to 59.0 MW (Fig. 10e).

annotType          startOffset  endOffset   annotId    Text         other
Quantity           36           39          1          One          {"mods":
                                                                    ["IsCount"]}
MeasuredEntity     40           45          2          Cycle        {"HasQuantity":
                                                                    " 1"}
Quantity           59           62          3          22%          {"unit": "%"}
MeasuredProperty   4            28          4          average      {"HasQuantity":
                                                       power        " 3"}
                                                       extracted
Measured Entity    36           45          5          one cycle    {"HasProperty":
                                                                    " 4"}
Qualifier          46           55          6          Increased    {"Qualifies": "
                                                                    4"}
...
```

*Figure 1- Sample initial result format for an example sentence text.*

### *Creating the Base Prompt*

With the experiment configuration established, the next step in preparation was the structure of the prompt and response strings. The prompt string submitted to GPT-3 consisted of two main parts, the base prompt (the representative few-shots of training data) and the specific MeasEval test case to process. The base prompt was the same for all MeasEval test sentences. Because this is few-shot learning, we could not use the entirety of the MeasEval training data. We did not sample the training data randomly. Instead, we took care to select examples for many of the permutations and combinations provided in the training data, such as singleton vs. range value quantities, numeric vs. textual quantity values, and missing vs. provided Unit and Property values.

The corresponding gold set annotations for these sampled paragraphs were identified and manually converted into our simplified, structured format consisting of sequences of labeled "Text" and annotation "Data" blocks. Each "Data" block consisted of one of more sequences of expected Quantity, Unit, Property, and Entity annotations providing the label and textual value for the label. Per recommendations from OpenAI, each structured paragraph in the prompt string was followed by an " <|endoftext|>" token (See Figure 2) Finally, after the final converted goldset annotation, a dangling "Text" label was added to trigger generation of the output.

```
Text: The averaged power extracted during one cycle increased
by 22% from 48.4 MW to 59.0 MW (Fig. 10e).

Data:
Quantity: one
MeasuredEntity: cycle

Quantity: 22%
MeasuredEntity: one cycle
MeasuredProperty: averaged power extracted
Unit: %

Quantity: 48.4 MW to 59.0 MW
MeasuredEntity: one cycle
MeasuredProperty: averaged power extracted
Unit: MW
<|endoftext|>
```

*Figure 2 - Sample prompt paragraph format*

The GPT-3 model has a request/response hard limit of 2049 "tokens". A "token" is described by OpenAI as approximately 4 characters of text. This token limit is applied to the sum of the prompt tokens and requested completion text size. When designing the base prompt, we had to compromise between providing sufficient training data in our "few-shot" submissions while preserving adequate space in the prompt to prevent truncation of the response from GPT-3. This resulted in a base prompt consisting of 7 example paragraphs containing 19 identified quantities with their associated attributes. The actual base prompt used in the experiment can be found in the appendix.

***Submitting to GPT-3***

The submission of the evaluation paragraphs was managed by a simple Python driver that leveraged the openai library. The complete set of paragraphs was broken into subsets of 25. This was done to automate the submission of paragraphs while limiting the consumption tokens in the event of a request exception from GPT-3. For each paragraph in each subset, the target paragraph was appended to the base prompt and submitted to the OpenAI API endpoint with the parameter shown in Figure 3:

```
params={
  "max_tokens":350,
  "temperature": 0.0,
  "engine": "davinci",
  "top_p":1.0,
}
```

*Figure 3 GPT-3 API Settings*

As previously noted, GPT-3 has a request/response hard limit of 2049 "tokens". When submitting our requests, we frequently had to dynamically reduce the max_tokens to return value in the generated response to keep the total length under the hard limit. This was due to variable length paragraphs, some extremely lengthy, being appended to the base prompt.

The results from each GPT-3 invocation were returned in a JSON structure from which we extracted the completion code and actual GPT-3 completion text. This data was the saved off for each paragraph in a separate file for further processing. The code for this driver routine and all the other processing referenced in this paper as well as the generated files can be found in our Github repo[9].

***Post Processing Results***

The formatted results from GPT-3 were not in the TSV format needed to evaluate them with the stand-alone MeasEval scoring program. In order to make them compliant, we needed to calculate span offsets within the paragraph, assign ids to the individual annotations, and construct the appropriate relationship annotations between the identified items. We did not attempt to recreate the Quantity modifiers or the Qualifier annotations. In order to transform our raw results, we ran each GPT-3 output file through a post-processing program. The one drawback to re-creating the dropped fields was a potential ambiguity when an identified text span might occur multiple times within the paragraph. Our approach for calculating the offsets was to use the first occurrence of the text, which could lead us to fail to differentiate subsequent occurrences of the string. Fifty-four annotations of different types across 38 paragraphs were impacted by this limitation.

Finally, there were also instances where GPT-3 would repeatedly output a series of annotations until the max token length for the response was reached for a paragraph. We applied a script provided by the MeasEval team to remove the duplicate annotation sets from our results so they would not skew the scores.

**Results**

We ran our post processed results through the MeasEval scoring program, which generated the following overall score for GPT-3. While the individual task papers for MeasEval have not yet been published, the scoring program did track the separate submissions. We see that our GPT-3 experiment placed very near the bottom (18th of 20).

| Precision | Recall | F-measure | Number of MeasEval Submissions | Effective MeasEval Rank |
|---|---|---|---|---|
| 0.402 | 0.157 | 0.226 | 20 | 18th |

*Table 4 - GPT-3 MeasEval Results*

We should note that this overall score could be artificially low as it was based on the performance for all the annotation types and modifiers where we had simplified our submission by dropping some of them. Fortunately, the scoring program provides the ability to report sub-scores on the results at the annotation type level. When looking at the four annotation types we actually generated and their associated Relations, the GPT-3 scores are shown in Figure 5:

| Annotation Type | Precision | Recall | F-measure | Number of MeasEval Submissions | Effective MeasEval Rank |
|---|---|---|---|---|---|
| Quantity | 0.631 | 0.395 | 0.486 | 20 | 19th |
| Unit | 0.612 | 0.421 | 0.499 | 20 | 16th |
| Entity | 0.229 | 0.127 | 0.164 | 14 | 10th |
| Property | 0.324 | 0.333 | 0.060 | 14 | 12th |
| HasQuantity | 0.130 | 0.739 | 0.094 | 14 | 11th |
| HasProperty | 0.172 | 0.015 | 0.028 | 13 | 11th |
| | | | | | |

*Figure 5 MeasEval SubTask Scores*

Overall GPT-3's performance was sporadic. It missed many seemingly simple Quantities in the evaluation set. In paragraph S0022459611006116-987, the sentence "The nitride fluorides show temperature independent paramagnetic behaviour between 5 and 300 K" generated a Quantity of "temperature independent paramagnetic behaviour" with a Unit annotation of the same value. More complex paragraphs frequently fared even worse. For example, paragraph S0032386113005454-2886 identified Quantities like: "localised shear-banding of the epoxy polymer", "internal cavitation of the S-CSR particles", "the S-CSR particles", and "the fracture energy" just to name a few.

Another aspect of the GPT-3 results we noticed, was that the model frequently had problems differentiating between Entity and Property annotations. In many cases combinations of the two types' spans would be grouped together in a single Entity or Property annotation. (e.g. evaluation paragraph S0032063313003218-5227 found an Entity of "Jovian upper atmospheric temperature" while the gold set had "temperature" identified as a separate Property annotation from a similar Entity annotation). The scoring algorithm attempts to find the best match in the gold set based on overlap of span offsets and the annotation type for a given annotation. In many of these cases, an annotation type/span would match a gold set annotation, albeit with a lower score due to the extra contents of the submission span, but there would be no related annotations for the associated Entity/Property. This may explain some of the poor performance GPT-3 exhibited for those annotation types and the related relationships sub-scores.

**Challenges Encountered**

*Prompt length restriction*

One of the major challenges we encountered was the 2049 maximum token length limit. The MeasEval task is rather complicated, encompassing 5 distinct annotation types, 4 related relationship types, and 11 possible Quantity modifiers. Almost all of these are optional so there is little uniformity amongst the paragraphs. An additional factor driving the prompt size is that MeasEval is defined over paragraphs, not sentences, and the paragraphs usually contain several sentences without measurements. In order to address this challenge, we simplified our test with GPT-3 to only have the system identify 4 of the 5 distinct annotation types, their associated Relations, and dropping the remainder of the items.

The OpenAI API enforces the token length restriction based on the sum of the size of the prompt and the requested max number of returned tokens. Should a request exceed the token limit, it automatically fails within the client. In order to circumvent this issue, at OpenAI's suggestion we leveraged the GPT-2 tokenizer from the transformers library to estimate the size of each prompt before submitting it. Even though there were some discrepancies between the token counts generated by both GPT versions, we were able to successfully dynamically update the max_tokens API parameter to successfully submit each paragraph. We should also note that OpenAI has identified this limitation and is considering adding a "truncate" option to the API in the future.

*Creative responses*

A more severe issue relates to the actual output from GPT-3. We had numerous cases where the model generated text spans that did not actually exist within the target paragraph. For example, one paragraph (S2213158213001253-2583) returned a Quantity of "1016 cm-3", a Unit of "cm-3" and an Entity of "injection level range". None of these text spans were actually present in the target paragraph. We have also seen cases where the generated text has changed the casing of words from the sentence and other instances where they are preserved. Finally, we have observed occurrences where returned spans are stitched together from non-contiguous spans in the paragraph.

On other occasions, we observed GPT-3 fall into a repetitive loop of output. For example, paragraph S0019103512003533-5211 generated the same sequence of annotations until the

max_token limit was encountered: Quantity: "10 keV electron energy flux", Unit: "keV", and Entity: "simulations R1-R18". In all, 26 of the 135 evaluation paragraphs were truncated for length by this issue.

*Response volatility*

The previous problems can, at least, be detected. But the most severe issue is that the responses generated by GPT-3 are very sensitive to the prompt that is provided to the model and there is no obvious way to predict what changes to a prompt will make the results better or worse. In our ad-hoc testing we observed slightly different prompts generating different outputs. We observed this to be true even when the difference was simply swapping the order of two examples in a prompt. Others, in independent research, have also recently reported this phenomenon as well[10][11]. This implies that the work of creating the few shot prompt is actually one of the key activities for leveraging GPT-3, and that getting a good generic prompt that will work across various inputs is not a trivial undertaking, if possible at all.

## Conclusion

While good for some tasks like generating human-like prose from a few words/sentences; simple translation tasks; and even potentially generating computer source code from a natural language description, GPT-3's ability to retain and identify factual information was weak. Additionally, in our opinion, inherent limiting factors such as prompt size restrictions as well as the model's literary license makes it ill-suited for complex, fact-based extraction tasks. We look forward to further progress in few-shot learning as a means of reducing the cost of creating training data, but shortcuts to tasks like MeasEval seem to be several years away.

## Acknowledgements

We would like to thank OpenAI for providing us access to their GPT-3 beta trial, their suggestions on improving performance, and their generous contribution of an additional set of credits to allow us to complete our experiment.

## Bibliography

**[1]** Tom Brown, Benjamin Mann, Nick Ryder, Melanie Subbiah, Jared D Kaplan, Prafulla Dhariwal, Arvind Neelakantan, Pranav Shyam, Girish Sastry, Amanda Askell, Sandhini Agarwal, Ariel HerbertVoss, Gretchen Krueger, Tom Henighan, Rewon Child, Aditya Ramesh, Daniel Ziegler, Jeffrey Wu, Clemens Winter, Chris Hesse, Mark Chen, Eric Sigler, Mateusz Litwin, Scott Gray, Benjamin Chess, Jack Clark, Christopher Berner, Sam McCandlish, Alec Radford, Ilya Sutskever, and Dario Amodei. 2020. *Language models are few-shot learners. In Advances in Neural Information Processing Systems*, volume 33, pages 1877–1901. Curran Associates, Inc.

[**2**] S. Rajbhandari, J. Rasley, O. Ruwase and Y. He, "ZeRO: Memory optimizations Toward Training Trillion Parameter Models," *SC20: International Conference for High Performance Computing, Networking, Storage and Analysis*, 2020, pp. 1-16, doi: 10.1109/SC41405.2020.00024.

[**3**] M. Onat Topal and Anil Bas and Imke van Heerden 2021 *Exploring Transformers in Natural Language Generation: GPT, BERT, and XLNet* arXiv:2102.08036 [cs.CL]

[**4**] Ke-Li Chiu, Rohan Alexander 2021 *Detecting Hate Speech with GPT-3* arXiv:2103.12407

[**5**] Jiachang Liu , Dinghan Shen , Yizhe Zhang , Bill Dolan , Lawrence Carin , Weizhu Chen *What Makes Good In-Context Examples for GPT-3* arXiv:2101.06804 [cs.CL]

[**6**] Xiao Liu and Yanan Zheng and Zhengxiao Du and Ming Ding and Yujie Qian and Zhilin Yang and Jie Tang 2021 *GPT Understands Too* arXiv:2103.10385 [cs.CL]

[**7**] Corey Harper, Jessica Cox, Curt Kohler, Antony Scerri, Ron Daniel Jr., and Paul Groth. 2021. SemEval 2021 task 8: MeasEval – extracting counts and measurements and their related contexts. In Proceedings of the Fifteenth Workshop on Semantic Evaluation (SemEval-2021), Bangkok, Thailand (online). Association for Computational Linguistics.

[**8**] Harper, Corey (2021) MeasEval Github Repo [Data] https://github.com/harperco/MeasEval

[**9**] Kohler, Curt (2021) MeasEval GPT-3 Utilities [Computer Software] https://github.com/elsevierlabs-os/measeval-gpt-3

[**10**] Yao Lu, Max Bartolo, Alastair Moore, Sebastian Riedel, and Pontus Stenetorp 2021 *Fantastically Ordered Prompts and Where to Find Them: Overcoming Few-Shot Prompt Order Sensitivity* arXiv:2104.08786 [cs.CL]

[**11**] Tony Z. Zhao and Eric Wallace and Shi Feng and Dan Klein and Sameer Singh *Calibrate Before Using: Improving Few-Shot Performance of Language Models arXiv:*2102.09690 [cs.CL]

## Appendix - Base prompt used in GPT-3 experiment

The actual prompt submitted to GPT-3 for each trial consisted of the following material, appended with one of the MeasEval test paragraphs, followed by “Data:”.

```
Text:
The particles also significantly toughened the epoxy polymer even at about −100 °C.

Data:
Quantity: about −100 °C
Unit: °C
Property: toughened
Entity: particles
<|endoftext|>
Text:
(I–M) Teratoma formation at 3 months after transplantation in the testes of SCID mice.

Data:
Quantity: 3 months
Unit: months
Property: after transplantation
Entity: Teratoma formation
<|endoftext|>
```

Text:
The hiPSCs cultured in spinner flasks for more than 10 passages not only could be remained pluripotent as indicated by in vitro and in vivo assays, but also could be efficiently induced toward mesodermal and hematopoietic differentiation.

Data:
Quantity: 10 passages
Unit: passages
Property: cultured
Entity: hiPSCs
<|endoftext|>
Text:
An additional uncertainty arises from saturation profiling, which discerns between pores filled with brine and pores filled with $CO_2$ above a poorly constrained level of saturation (Arts et al., 2004). It follows from the commonly used Gassmann equation (Gassmann, 1951) that: (a) very low saturations of less than a few percent are undetectable; (b) a strong correlation emerges as the gas saturation increases from a few percent to around thirty percent; and (c) for saturations much above thirty percent, it is difficult to distinguish between moderate and high saturations. It follows that the 80% gas saturation that is commonly assumed for the Sleipner plume, while reasonable (Chadwick et al., 2005; Bickle et al., 2007), remains uncertain (Lumley, 2008). If the 80% assumption represents a reasonable upper limit to the mean saturation for the plume, the lower limit could be as low as 40%, halving mass balance estimates premised on the widely assumed high-saturation value.

Data:
Quantity: a few percent to around thirty percent
Unit: percent
Entity:  gas saturation

Quantity: above thirty percent
Unit: percent
Entity: saturations

Quantity: 80%
Unit: %
Property: gas saturation
Entity: Sleipner plume

Quantity: 80%
Unit: %
Property: upper limit
Entity: mean saturation for the plume

Quantity: 40%
Property: lower limit
Entity: mean saturation for the plume
<|endoftext|>
Text:
Quantitative imaging analysis revealed that approximately 78 ± 0.6% of cells were Sox17-positive on D5 and approximately 88 ± 1.1% of cells were AFP-positive on D13 and approximately 29 ± 0.9% of cells were ALB-positive on D20 (Fig. 3D). The ALB secretion in the media of differentiated FAP-specific iPS cells on D20 was approximately 20 µg/ml (Fig. 3E). Moreover, these D20 differentiated FAP-specific iPS cells were also periodic acid-Schiff (PAS)-positive, indicating cytoplasmic glycogen storage (Fig. 3F). These results clearly indicated that FAP-specific iPS cells had the potential to differentiate into hepatocyte-like cells.

Data:
Quantity: approximately 78 ± 0.6%
Unit: %
Property: Sox17-positive on D5
Entity: cells

Quantity: approximately 88 ± 1.1%
Unit: %
Property: AFP-positive on D13
Entity: cells

Quantity: approximately 29 ± 0.9%
Unit: %
Property: ALB-positive on D20
Entity: cells

Quantity:  approximately 20 µg/ml
Unit: µg/ml
Entity:  1ALB secretion
<|endoftext|>
Text:
In vivo incorporation of 35S-methionine/cysteine into mitochondrially encoded proteins was visualised by separation of cell lysate (50 µg) through SDS-PAGE, exposure of the dried gel to a PhosphorImage screen, followed by Storm and ImageQuant analysis (upper panel).

Data:
Quantity: 50 µg
Unit: µg
Entity:  cell lysate
<|endoftext|>
Text:
There is a red-shift in the solid state CD spectrum of GMP compared with its liquid CD spectrum. The negative band about the conformation of the sugar moiety has nearly disappeared. In solid-state, mutarotation is more difficult and the negative band at the 260 nm nearby can be detected which indicates that the GMP ligand is mainly β-anomers. In the spectrum of complex 1, the weak negative band at 215–225 nm indicates that the GMP exists as l-ribo [5], which is induced by the strong π–π and hydrogen bonding interaction. For solid samples, CD spectroscopy is highly sensitive to even a very small distortion from planarity of the aromatic chromophore [12]. The strong negative CE centered at 297 nm (θ ≈ – 10 mdeg) is due to the excitation coupling of π → π* transitions of the aromatic chromophores, including the intra- and intermolecular coupling of the guanine chromophores [13], which is consistent with the single crystal structure.

Data:
Quantity: 260 nm
Unit: nm
Property: negative band
Entity: solid state CD spectrum of GMP

Quantity: 215–225 nm
Unit: nm
Property: negative band
Entity: spectrum of complex 1

Quantity: 297 nm
Unit: nm
Entity: strong negative CE

Quantity: ≈ – 10 mdeg
Unit: mdeg
Entity: θ
<|endoftext|>
Text: Serial passaging of two different hiPSC lines in the spinner flasks using the E8 medium preserved their normal karyotype and expression of undifferentiated state markers of TRA-1-60, SSEA4, OCT4, and NANOG.

Quantity: two
Entity: hiPSC lines
<|endoftext|>
Text: The injection level range studied varied with the lifetime of the sample, but was usually in the range 1013 cm−3 to 1016 cm−3.

Data:
Quantity: range 1013 cm−3 to 1016 cm−3
Unit: cm−3
Entity: injection level range
<|endoftext|>
Text: